\documentclass{article}

\usepackage[preprint,nonatbib]{neurips_2018}

\usepackage[utf8]{inputenc} 
\usepackage[T1]{fontenc}    
\usepackage{hyperref}       
\usepackage{url}            
\usepackage{booktabs}       
\usepackage{amsfonts}       
\usepackage{nicefrac}       
\usepackage{microtype}      
\usepackage{graphicx} 
\usepackage{amsmath} 
\usepackage{mathdots} 
\usepackage[title]{appendix}
\usepackage[colorinlistoftodos]{todonotes}
\usepackage{todonotes} 
\usepackage{multirow}
\usepackage{algorithm,algorithmicx,algpseudocode}

\newcommand{\grad}{\nabla}
\newcommand{\bx}{\mathbf{x}}
\newcommand{\bz}{\mathbf{z}}
\newcommand{\bepsilon}{{\boldsymbol{\epsilon}}}
\newcommand{\bzero}{\mathbf{0}}
\newcommand{\bI}{\mathbf{I}}

\title{Advancing Diffusion Models: Alias-Free Resampling and Enhanced Rotational Equivariance}

\author{%
  Md Fahim~Anjum\thanks{Alternate email: dr.fahim.anjum@gmail.com} \\
  Department of Neurology\\
  University of California San Francisco\\
  San Francisco, CA 94143 \\
  \texttt{fahim.anjum@ucsf.edu} \\
}

\begin{document}

\maketitle

\begin{abstract}
 Recent advances in image generation, particularly via diffusion models, have led to impressive improvements in image synthesis quality. Despite this, diffusion models are still challenged by model-induced artifacts and limited stability in image fidelity. In this work, we hypothesize that the primary cause of this issue is the improper resampling operation that introduces aliasing in the diffusion model and a careful alias-free resampling dictated by image processing theory can improve the model's performance in image synthesis. We propose the integration of alias-free resampling layers into the UNet architecture of diffusion models without adding extra trainable parameters, thereby maintaining computational efficiency. We then assess whether these theory-driven modifications enhance image quality and rotational equivariance. Our experimental results on benchmark datasets, including CIFAR-10, MNIST, and MNIST-M, reveal consistent gains in image quality, particularly in terms of FID and KID scores. Furthermore, we propose a modified diffusion process that enables user-controlled rotation of generated images without requiring additional training. Our findings highlight the potential of theory-driven enhancements such as alias-free resampling in generative models to improve image quality while maintaining model efficiency and pioneer future research directions to incorporate them into video-generating diffusion models, enabling deeper exploration of the applications of alias-free resampling in generative modeling.
\end{abstract}

\section{Introduction}
Recent advancements in generative modeling, particularly in diffusion models \cite{ddpm}, have pushed the boundaries of what is possible in high-quality image synthesis. Among these, the Stable Diffusion model has gained prominence for its ability to generate realistic images by iteratively refining noise into coherent visual outputs \cite{stablediff3}. Despite its success, there remains a challenge in further enhancing the model's performance, particularly in terms of stability and image fidelity \cite{chen2024opportunities}.

In this paper, we hypothesize that the existing resampling operations (upsampling/downsampling) in the architecture of current diffusion models introduce aliasing which leads to a reduction of image quality. We also propose that proper theory-driven alias-free resampling can improve the model's performance in image synthesis. Improving the performance of image synthesis via alias-free resampling techniques has recently been explored in generative adversarial networks (GANs). Indeed, StyleGAN3 \cite{stylegan3}, the latest iteration in the StyleGAN series, has demonstrated significant improvements over its predecessors by incorporating carefully designed  alias-free resampling layers via anti-aliasing filtering techniques that prevent high-frequency artifacts and improve the overall visual coherence of generated images. Earlier versions of StyleGAN networks, including StyleGAN2 \cite{stylegan2}, fail to rigorously implement the alias-free resampling during the up or downsampling stages and this careless signal processing can be a root cause of aliasing in the generator network \cite{stylegan3}. Their advancements have shown that even small architectural modifications, when grounded in image processing principles, can lead to substantial gains in model performance. Alias-free resampling ensures that when images are upsampled or transformed between scales, no high-frequency components are artificially introduced, which could corrupt the learned texture or details. Thus, incorporating these principles allows the model to avoid aliasing artifacts and enhance its rotational equivariance and overall output fidelity.

In addition to GAN networks, we hypothesize that the principle of alias-free resampling is especially crucial for diffusion models, where sequential resampling occurs across multiple scales, making it essential to preserve fine details and avoid introducing spurious artifacts. However, to date, proper integration of alias-free resampling into diffusion models remains largely unexplored. Current diffusion models typically apply standard downsampling and upsampling operations \cite{ddpm}, which can introduce aliasing and degrade the quality of generated images, especially at finer scales. By integrating alias-free resampling techniques, diffusion models could achieve more stable and artifact-free outputs, enhancing both image fidelity and rotational consistency across generated samples.

This work investigates the theory-driven integration of alias-free resampling techniques into the UNet structure of Diffusion models. Importantly, our approach focuses on enhancing model performance without introducing any new trainable parameters, thereby maintaining the model's efficiency and simplicity. By strategically incorporating alias-free resampling layers, we aim to leverage the principles of image processing to improve the stability and output quality of the diffusion model. Furthermore, we propose a modified diffusion process to incorporate user-controlled rotation of the generated image without any additional training.

Our experiments show that incorporating our proposed modifications significantly enhances the image quality across various configurations of the UNet structure. Specifically, our modified configurations outperformed the standard UNet model on benchmark datasets such as MNIST, CIFAR-10, and MNIST-M \cite{cifar10,mnist,mnistm}. This highlights the potential of using appropriate alias-free resampling layers, as guided by image processing principles, to achieve better results in generative models. Our modified diffusion process also showed promising results in rotational consistency despite being trained on images without rotations. The key takeaway from our findings is that careful architectural re-design governed by signal and image processing theories can enhance model performance without the need for additional trainable parameters, thereby offering a path forward for further innovations in generative modeling.

\section{Theoretical Foundations}
\subsection{Principles of Alias-Free Resampling}

Alias-free resampling is a critical signal processing technique for mitigating aliasing in 1D signals and 2D images, ensuring that high-frequency details are faithfully represented without introducing unwanted artifacts. For models like diffusion-based architectures, aliasing can occur due to improper downsampling and upsampling operations or nonlinearities like ReLU, causing artifacts that disrupt the generation of high-quality images. Alias-free resampling addresses this by adhering to the Shannon-Nyquist sampling theorem \cite{shannon1949communication}, a fundamental theoretical principle that provides the necessary conditions for sampling a signal or image. At its core, this theorem states that if we uniformly sample a signal, the sampling rate has to be at least twice the highest frequency of the signal's bandwidth.  Conversely, the bandwidth of a discrete signal has to be half the sampling rate. If the sampled signal contains frequencies beyond this limit, aliasing occurs, meaning that high-frequency components will overlap with lower frequencies, distorting the reconstructed signal. Suppose $x(t)$ is a 1D signal that was uniformly sampled at a rate \( s \) to obtain a discrete signal \( x[n] \). Then, the Shannon-Nyquist sampling theorem dictates that the frequency bandwidth of $x[n]$  must lie within the Nyquist limit, which is half the sampling rate, \( f_{\text{Nyquist}} = \nicefrac{s}{2} \) \cite{shannon1949communication}. Therefore, before resampling a signal, it is critical to apply a low-pass filter (also known as an anti-aliasing filter) with a frequency cutoff up to $f_{\text{Nyquist}}$ to stop the aliasing. The principle is identical for 2D signals such as images.

\subsection{The Role of Alias-Free Resampling in Nonlinear Transformations}
Nonlinear operations like GeLU or ReLU in the continuous domain introduce sudden fluctuations causing arbitrarily high frequencies that cannot be represented in the sampled output and a natural solution is to eliminate the offending high-frequency content by applying an ideal low-pass filter. However, diffusion networks utilize discrete domain data where point-wise nonlinearity is utilized which does not commute with fractional transformations (such as rotation). Therefore, to temporarily approximate a continuous representation, we utilize a proper $2\times$ alias-free upsampling, apply the nonlinearity in the higher resolution and finally use a $2\times$ alias-free downsampling equipped with low-pass anti-aliasing filter for returning to the original discrete space.

\subsection{Designing Anti-Aliasing Filters}
\begin{figure}[t]
	\centering
	\includegraphics[width=14cm,height=11.5cm,clip,keepaspectratio]{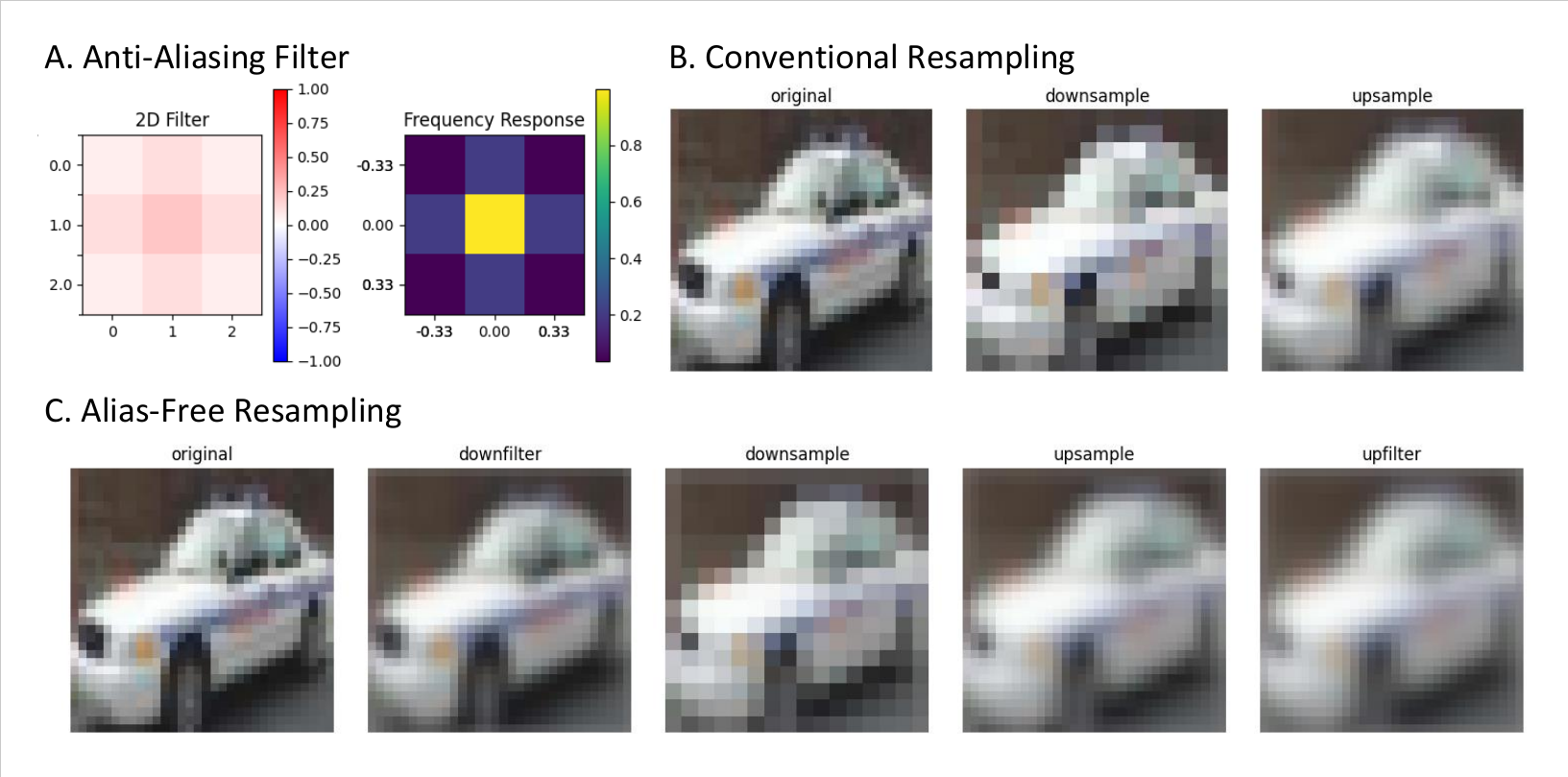}
	\caption{Alias-free resampling via anti-aliasing low-pass filters. Panel A shows a $3\times 3$ anti-aliasing filter and its frequency response with Kaiser window ($\beta = 1$). Panel B shows conventional resampling operations ($2\times$ downsampling followed by $2\times$ upsampling) and panel C shows alias-free resampling operations with anti-aliasing filters (downfilter and upfilter steps) and upsampling with interleaved zeros (upsample step). }
	\label{Fig0}
\end{figure}
As discussed in the previous section, alias-free resampling involves applying a low-pass filter to the signal to ensure that no aliasing occurs (Figure \ref{Fig0}). In particular, given a discrete 2D signal \( x[n_1, n_2] \) sampled on a regular grid with spacing \( 1/s \), alias-free resampling applies a low-pass filter with a cutoff frequency \( f_c \leq f_{\text{Nyquist}} \) (or \( \omega_c \leq \nicefrac{\pi}{2} \) ), ensuring that no frequencies above the Nyquist limit are included in the resampled signal. The 2D circularly symmetric low-pass filter has an impulse response \( h[n_1, n_2] \) given by:
\begin{equation}\label{eq_h}
	h[n_1, n_2] = \frac{\omega_c}{2\pi \sqrt{n_1^2 + n_2^2}} J_1 \left( \omega_c \sqrt{n_1^2 + n_2^2} \right)
\end{equation}
where \( J_1 \) is the Bessel function of the first kind and of the first order \cite{lim1990two}. At the center point, \( n_1 = n_2 = 0 \), the limiting value is used: \( h[0, 0] = \nicefrac{\omega_c^2}{4\pi} \).
The impulse response in (\ref{eq_h}) is also known as the Jinc function, analogous to the sinc function in 1D case, and is defined as:
\begin{equation}\label{eq_j}
\text{jinc}(x) = \frac{J_1(x)}{x}
\end{equation}
Using this, the impulse response can be rewritten as:
\begin{equation}\label{eq_h2}
h[n_1, n_2] = \frac{\omega_c^2}{2\pi} \text{jinc}\left( \omega_c \rho \right), \quad \rho = \sqrt{n_1^2 + n_2^2}
\end{equation}

This circular low-pass filter ensures that only frequencies below the cutoff \( \omega_c \) are preserved, effectively eliminating aliasing by attenuating higher frequencies (Figure \ref{Fig0}). By convolving this filter with the sampled signal, alias-free resampling is achieved, preventing unwanted artifacts and ensuring fidelity. The filtered image is obtained by convolving $x$ with $h$:
\begin{equation}\label{eq_filter}
x_{\text{filtered}}[n_1, n_2] = \sum_{i, j} h[i, j] \cdot x[n_1 - i, n_2 - j]
\end{equation}
An ideal filter that completely eliminates frequencies above $\omega_c$ has an infinite impulse response, making it impractical due to implementation inefficiency. To address these issues, the filter is typically truncated using the window method, where a window function limits the spatial extent of the filter resulting in a practical approximation. While different window functions balance trade-offs between frequency response and spatial extent, in this work we used the Kaiser window\cite{kaiser1974}, which provides control over this trade-off through its shape parameter $\beta$ and spatial extent $L$, and is defined using the zeroth-order modified Bessel function $I_0$:
\begin{equation}\label{eq_wk}
w_K(n) = 
\begin{cases} 
	\frac{I_0\left( \beta \sqrt{1 - \left( \frac{2n}{L} \right)^2} \right)}{I_0(\beta)}, & \text{if } |n| \leq \frac{L}{2}, \\
	0, & \text{if } |n| > \frac{L}{2},
\end{cases}
\end{equation}
Finally, we can also normalize the filter such that $\sum_{i, j} h[i, j]=1$. This makes sure that the total scaling is constant. Thus, we have the following parameters for the anti-aliasing resampling process:
\begin{enumerate}
	\item Filter cutoff ($\omega_c$):  We kept this fixed ($\omega_c=\nicefrac{\pi}{2}$).
	\item Kernel length: We keep the kernel length of our filters to be fixed at $3 \times 3$.
	\item Kaiser shape ($\beta$): We varied this between 0 (no effective Kaiser window), 1, and 2.
	\item Normalization: We tested both normalized and un-normalized kernels.
\end{enumerate}

\section{Architectural Revisions in Diffusion Model}\label{methods}
\subsection{Baseline Architecture (Config A)}
We use a classical unconditional diffusion model with standard noising and denoising steps as our baseline where we follow Algorithm \ref{alg:training} from \cite{ddpm} for training. The baseline architecture is based on a UNet encoder-decoder structure with skip connections, where the input image is progressively $2\times$ downsampled to capture high-level features and subsequently $2\times$ upsampled to recover fine details. 

\algrenewcommand\algorithmicindent{0.5em}%
\begin{figure}[t]
	\begin{minipage}[t]{\textwidth}
		\begin{algorithm}[H]
			\caption{Training \cite{ddpm}} \label{alg:training}
			\small
			\begin{algorithmic}[1]
				\Repeat
				\State $\bx_0 \sim q(\bx_0)$
				\State $t \sim \mathrm{Uniform}(\{1, \dotsc, T\})$
				\State $\bepsilon\sim\mathcal{N}(\bzero,\bI)$
				\State Take gradient descent step on
				\Statex $\qquad \grad_\theta \left\| \bepsilon - \bepsilon_\theta(\sqrt{\bar\alpha_t} \bx_0 + \sqrt{1-\bar\alpha_t}\bepsilon, t) \right\|^2$
				\Until{converged}
			\end{algorithmic}
		\end{algorithm}
	\end{minipage}
	$\bx_0, q(.),\bepsilon_\theta(.)$ and $\bar\alpha_t$ are defined in \cite{ddpm}
	\vspace{-1em}
\end{figure}

\begin{figure}[t]
	\centering
	\includegraphics[width=15.5cm,height=14.5cm,clip,keepaspectratio]{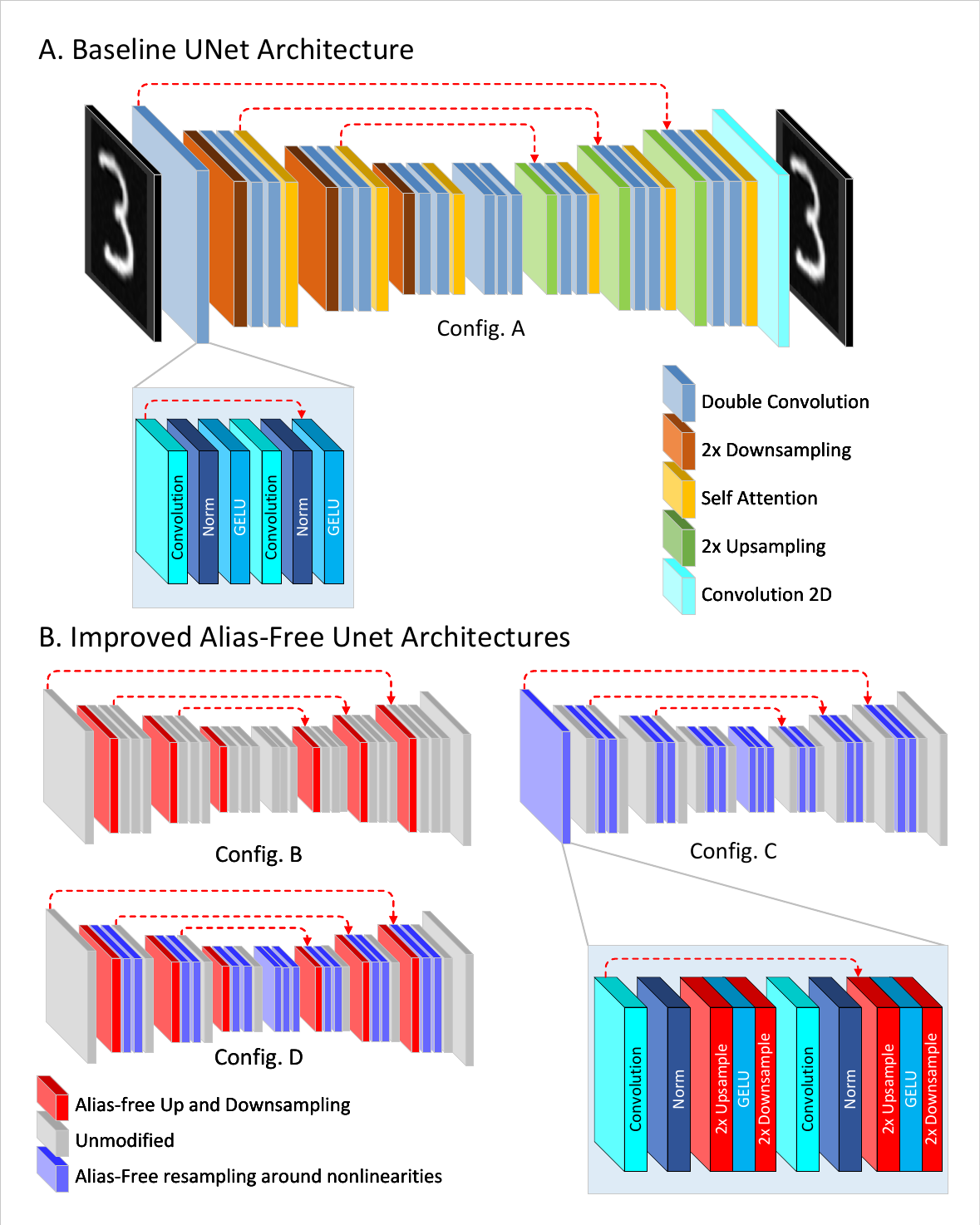}
	\caption{Overview of the conventional baseline UNet (Panel A) and our architectural revisions (Panel B) of the baseline UNet in diffusion models.}
	\label{Fig1}
\end{figure}

The input first passes through a series of convolutional layers. Conventional downsampling is achieved through max pooling, which selects the maximum value in non-overlapping regions, reducing the spatial resolution by half. Upsampling is performed through conventional bilinear interpolation, which smoothly increases the resolution by averaging the neighboring pixel values. Additionally, an alignment step ensures that the corners of the input and output grids are matched, preserving the spatial consistency across layers. The bottleneck section consists of several convolutional layers. Skip connections pass intermediate features from the downsampling stages to their corresponding upsampling stages, ensuring that important spatial information is retained.
Additionally, self-attention layers are applied in both the downsampling and upsampling stages to refine the feature maps by considering the global context. We denote this baseline architecture as Config A (Figure \ref{Fig1}).
 
\subsection{Alias-Free Resampling (Config B)}
First, we revise our baseline architecture (Config A) by replacing the up and downsampling layers, which do not guarantee alias-free outputs, with their alias-free versions respectively. In particular, the downsampling layers first apply low-pass anti-aliasing filters to the data as described in (\ref{eq_filter}), and then reduce the sampling rate by uniformly removing interleaving samples (Figure \ref{Fig0}). During upsampling, zeros are interleaved with the data to increase the sampling rate, followed by passing the result through a low-pass filter to remove unwanted high-frequency components (Figure \ref{Fig0}). These steps are grounded in classical resampling theory from image processing. We denote this modified version as Config. B (Figure \ref{Fig1}), which significantly improves the quality of resampling, reducing aliasing artifacts and enhancing output fidelity.

\subsection{Enhanced Nonlinearities via Alias-Free Resampling (Config C)}
Next, we shift our attention to the nonlinear components of our baseline architecture (Config. A) and revise it by introducing $2\times$ alias-free upsampling before the nonlinear ReLU or GeLU operations and $2\times$ alias-free downsampling afterward to retain the original sampling rate. These adjustments aim to mitigate aliasing introduced by the nonlinear operations while preserving high-frequency details. Importantly, the existing upsampling and downsampling layers in the network remain unmodified in this configuration (Config. C; Figure \ref{Fig1}), as we only inject alias-free resampling layers around the nonlinear operations.

\subsection{Combining Alias-Free Resampling and Nonlinear Enhancements (Config D)}
Here, we combine Configurations B and C by replacing the upsampling and downsampling layers in Configuration C with their alias-free counterparts from Configuration B. This ensures that both the nonlinear operations and the standard resampling processes in the network are alias-free, effectively reducing artifacts and improving image fidelity across all stages. We denote this revised architecture as Config. D (Figure \ref{Fig1}).

\subsection{Improving Rotational Consistency}
\begin{figure}[t]
	\centering
	\includegraphics[width=14cm,height=13.5cm,clip,keepaspectratio]{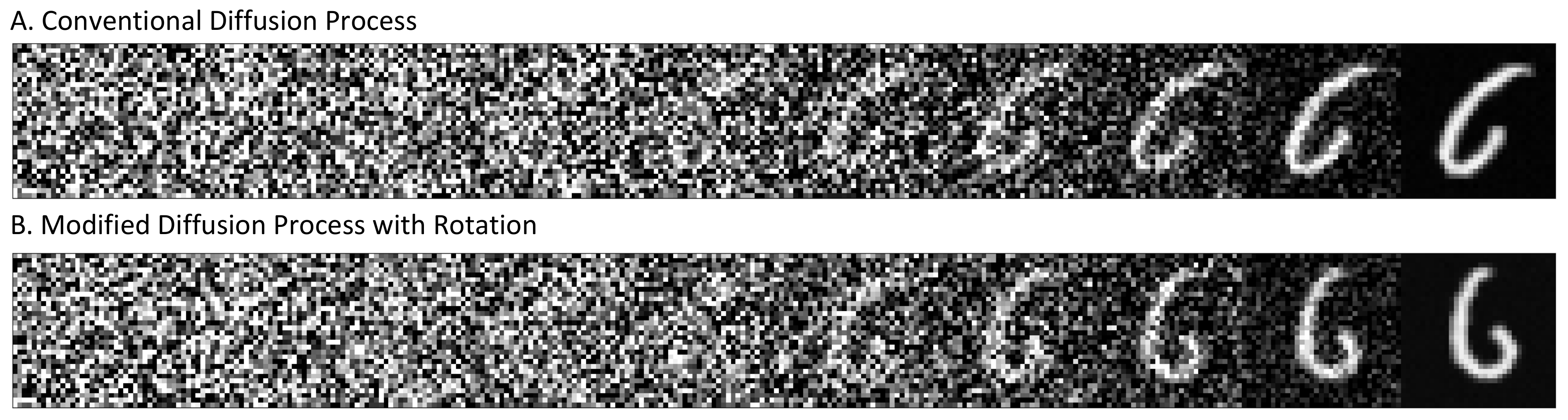}
	\caption{Improving rotational consistency with modified diffusion process: Panel A shows the classical diffusion process while Panel B illustrates our proposed modified diffusion process, achieving counter-clockwise rotation.}
	\label{Fig1.1}
\end{figure}

Lastly, we revise the classical diffusion process (Algorithm \ref{alg:sampling}) to incorporate controlled rotation during image generation. The core idea is to progressively distribute the target rotation over the time steps (Figure \ref{Fig1.1}). At each time step, the image is rotated by a small, constant angle, ensuring that as the diffusion progresses, the image gradually rotates towards a target orientation (Algorithm \ref{alg:sampling2}). This modification allows the model to generate images with user-defined rotational transformations while maintaining coherence throughout the generative process. 

In particular, given the input matrix to be rotated \(\bx\) and the rotation angle \(\phi\), the rotation function can be represented as, \( \text{Rotate} \left( \bx, \phi \right) \) which performs an affine transformation on \(\bx\), effectively rotating it around its center by \(\theta\). This process begins by determining the center of the image, which serves as the pivot point for the rotation. The positions of all pixels are then adjusted according to \(\phi\), recalculating their coordinates to reflect the desired rotation. This involves translating the coordinates so that the rotation occurs around the image center and then translating them back to their original location. Note that there are parts of the image that will have to be extrapolated due to the rotation operation. Finally, the rotation is applied at each time step. Therefore, if the desired rotation is  \(\phi\), rotation at time step $t$ is, \(\phi_t=\nicefrac{\phi}{T}\). 

\algrenewcommand\algorithmicindent{0.5em}%
\begin{figure}[t]
	\begin{minipage}[t]{0.495\textwidth}
		\begin{algorithm}[H]
			\caption{ Classical Sampling \cite{ddpm}} \label{alg:sampling}
			\small
			\begin{algorithmic}[1]
				\vspace{.04in}
				\State $\bx_T \sim \mathcal{N}(\bzero, \bI)$
				\For{$t=T, \dotsc, 1$}
				\State $\bz \sim \mathcal{N}(\bzero, \bI)$ if $t > 1$, else $\bz = \bzero$
				\State $\bx_{t-1} = \frac{1}{\sqrt{\alpha_t}}\left( \bx_t - \frac{1-\alpha_t}{\sqrt{1-\bar\alpha_t}} \bepsilon_\theta(\bx_t, t) \right) + \sigma_t \bz$
				\EndFor
				\State \textbf{return} $\bx_0$
				\vspace{.2in}
			\end{algorithmic}
		\end{algorithm}
	\end{minipage}
	\hfill
	\begin{minipage}[t]{0.495\textwidth}
		\begin{algorithm}[H]
			\caption{Sampling with Rotation} \label{alg:sampling2}
			\small
			\begin{algorithmic}[1]
				\vspace{.04in}
				\State $\bx_T \sim \mathcal{N}(\bzero, \bI)$
				\For{$t=T, \dotsc, 1$}
				\State $\bz \sim \mathcal{N}(\bzero, \bI)$ if $t > 1$, else $\bz = \bzero$
				\State $\bx_{t-1} = \frac{1}{\sqrt{\alpha_t}}\left( \bx_t - \frac{1-\alpha_t}{\sqrt{1-\bar\alpha_t}} \bepsilon_\theta(\bx_t, t) \right) + \sigma_t \bz$
				\State $\bx_{t-1} = \text{Rotate}(\bx_{t-1}, \frac{\phi}{T})$
				\EndFor
				\State \textbf{return} $\bx_0$
				\vspace{.04in}
			\end{algorithmic}
		\end{algorithm}
	\end{minipage}
	$\bx_0, q(.),\bepsilon_\theta(.), \alpha_t, \sigma_t$ and $\bar\alpha_t$ are defined in \cite{ddpm}
	\vspace{-1em}
\end{figure}
  
\subsection{Configuration Naming Scheme}
In addition to the Alphabets (A-D) we use for denoting our revised UNet architecture, we also use two parameters during the naming of our models. These are the Kaiser $\beta$ value and whether or not the kernel was normalized. Specifically, the $\beta$ value is added after the configuration alphabet, indicating the specific Kaiser window parameter used. If the kernel was normalized, we append the letter 'N' at the end. For example, a model denoted as "Config B-2N" refers to Configuration B, with a Kaiser $\beta$ value of 2 and a normalized filter kernel. These naming schemes are summarized in Table \ref{tab:name} with examples.

\begin{table}[t]
	\caption{Summary of the architectural revisions and naming convention.}
	\label{tab:name}
	\centering
	\begin{tabular}{llcc}
		\toprule
		\multirow{2}{*}{\textbf{Name}} & \multirow{2}{*}{\textbf{Architecture Details}} & \multicolumn{2}{c}{\textbf{Filter Properties}} \\
		\cmidrule(l){3-4}
		&  & \textbf{Normalized} & \textbf{Kaiser $\beta$} \\  
		\midrule
		Config. A & Baseline & - & - \\
		\midrule
		Config. B-0 & A + alias-free up and downsampling & No & 0 \\
		\midrule
		Config. C-1N & A + alias-free resampling around nonlinearities & Yes & 1 \\
		\midrule
		Config. D-2 & A + alias-free resampling around nonlinearities & No & 2 \\
		& \quad+ alias-free up and downsampling & & \\
		\bottomrule
	\end{tabular}
\end{table}

\section{Experiments}\label{sec:exp}
\subsection{Datasets and Experimental Setup}

To train and evaluate our model, we used three benchmark image datasets \cite{cifar10,mnist,mnistm}, each pre-processed to ensure consistency in input dimensions and pixel intensity normalization across experiments. In particular, for each dataset, we applied standard normalization across all channels by centering the pixel values around zero with a range of $[-1, 1]$. All images were resized to $32 \times 32$ pixels.

\subsubsection{CIFAR-10} 
CIFAR-10 dataset consists of 60,000 $32 \times 32$ color images in 10 classes, with 6,000 images per class \cite{cifar10}. We utilized a subset of 10,000 images (test set of the original dataset) from the CIFAR-10 dataset with 10 classes (1,000 images from each class). Each sample is a 3-channel RGB color image with a native resolution of $32 \times 32$ pixels. 
	
\subsubsection{MNIST} 
MNIST dataset of handwritten digit images consists of 60,000 single-channel grayscale samples \cite{mnist}. We used a subset of the MNIST dataset containing 19,999 samples, which was available in the Google Colab environment. Each image was initially $28 \times 28$ pixels which was resized to $32 \times 32$ pixels.
	
\subsubsection{MNIST-M} 
We also utilized MNIST-M, a variation of MNIST with added background textures \cite{mnistm}. We used 6,000 randomly selected samples from the original 60,000-image set. Each image is a 3-channel RBG color image. The native image dimensions of $28 \times 28$ pixels were resized to $32 \times 32$ pixels.

\subsection{Evaluation Metrics}
To benchmark the performance of the generative models, we utilized several metrics: Inception Score (IS), Fréchet Inception Distance (FID), and Kernel Inception Distance (KID). IS measures the diversity and quality of generated samples, with higher scores indicating better performance. FID score quantifies the difference between the generated and real data distributions by comparing their feature representations, where lower values correspond to more realistic samples while KID is a variation of FID that uses the squared maximum mean discrepancy between samples, providing an unbiased comparison, with lower values indicating better performance. These metrics were computed across all datasets to provide a comprehensive evaluation of the model's performance.
 
\subsection{Training Configurations}
We conducted unconditional training of the diffusion models, with a learning rate of $3 \times 10^{-3}$ and a batch size of 16 over 100 epochs. The model was trained with 1000 noise steps, where the noise schedule followed a linear progression, starting from an initial $\beta_{\text{noise}}$ value of $1 \times 10^{-4}$ and increasing to a final value of 0.02. No validation was performed during training. The loss function was based on the mean squared error between the predicted noise and the true Gaussian noise added to the data at each time step. This allowed the model to learn how to reverse the diffusion process by minimizing the error in noise prediction.

\section{Results}\label{sec:results}
\subsection{Standard Image Synthesis Performance}

In this section, we present the quantitative results across the CIFAR-10, MNIST-M, and MNIST datasets, comparing our modified configurations (Config B-D) against the baseline architecture (Config A).

\begin{table*}[t]
	\caption{Performance comparison on CIFAR-10, MNIST-M, and MNIST datasets.}
	\hfill
	\centering
	\label{table_perf}
	\begin{tabular}{|l||ccc||ccc||ccc|}
		\hline
		\multirow{2}{*}{\textbf{Configuration}} & \multicolumn{3}{c||}{\textbf{CIFAR-10}}   & \multicolumn{3}{c||}{\textbf{MNIST-M}}  & \multicolumn{3}{c|}{\textbf{MNIST}}     \\ 
		& IS$\uparrow$    & FID$\downarrow$    & KID*$\downarrow$ & IS$\uparrow$    & FID$\downarrow$    & KID*$\downarrow$ & IS$\uparrow$    & FID$\downarrow$    & KID*$\downarrow$  \\
		\hline
		\hline
		A (Baseline)     & 4.54  & 98.77  & 5.97  & 3.76   & 85.00  & 6.23  & 1.98  & \textbf{9.61}   & \textbf{0.47}  \\
		\hline
		B-0    & \textbf{4.71}  & 94.23  & \textbf{5.44}  & 3.39   & 93.81  & 7.37  & 1.99  & 10.23  & 0.58  \\
		C-0    & 3.75  & 129.42 & 7.92  & 3.11   & 124.10 & 9.43  & 1.94  & 14.07  & 0.96  \\
		D-0    & 4.33  & 97.44  & 6.67  & 3.33   & 98.16  & 7.56  & 1.94  & 14.37  & 1.01  \\
		\hline
		B-1    & 4.63  & 121.45 & 6.90  & 3.40   & 94.11  & 7.40  & 1.97  & 11.00  & 0.64  \\
		C-1    & 3.56  & 138.88 & 10.47 & 3.48   & 124.78 & 7.86  & 1.97  & 14.76  & 1.05  \\
		D-1    & 4.32  & 108.06 & 7.42  & 3.44   & 114.27 & 8.35  & 1.98  & 16.08  & 1.12  \\
		\hline
		B-1N   & 4.63  & 125.71 & 6.64  & 3.71   & 100.91 & 7.53  & 1.97  & 11.62  & 0.72  \\
		C-1N   & 3.99  & 107.37 & 6.96  & 3.69   & 144.41 & 9.69  & 1.96  & 15.95  & 1.23  \\
		D-1N   & 4.51  & \textbf{90.21}  & 5.54  & 3.68   & 108.14 & 7.65  & 1.96  & 14.25  & 0.97  \\
		\hline		
		B-2N   & 4.34  & 109.96 & 7.65  & \textbf{4.14}   & 88.05  & 5.47  & \textbf{2.00}  & 12.78  & 0.87  \\
		C-2N   & 4.34  & 95.11  & 6.70  & 4.01   & 101.59 & 6.78  & 1.97  & 16.73  & 1.29  \\
		D-2N   & 4.50  & 102.28 & 6.81  & 3.99   & \textbf{82.46}  &\textbf{ 5.35}  & 1.99  & 11.19  & 0.71  \\
		\hline
		\multicolumn{10}{l}{\footnotesize IS: Inception Score, FID: Frechet Inception Distance, KID: Kernel Inception Distance* ($\times$ 100).} \\
	\end{tabular}

\end{table*}

\subsubsection{CIFAR-10}

\begin{figure}[t]
	\centering
	\includegraphics[width=14cm,height=13cm,clip,keepaspectratio]{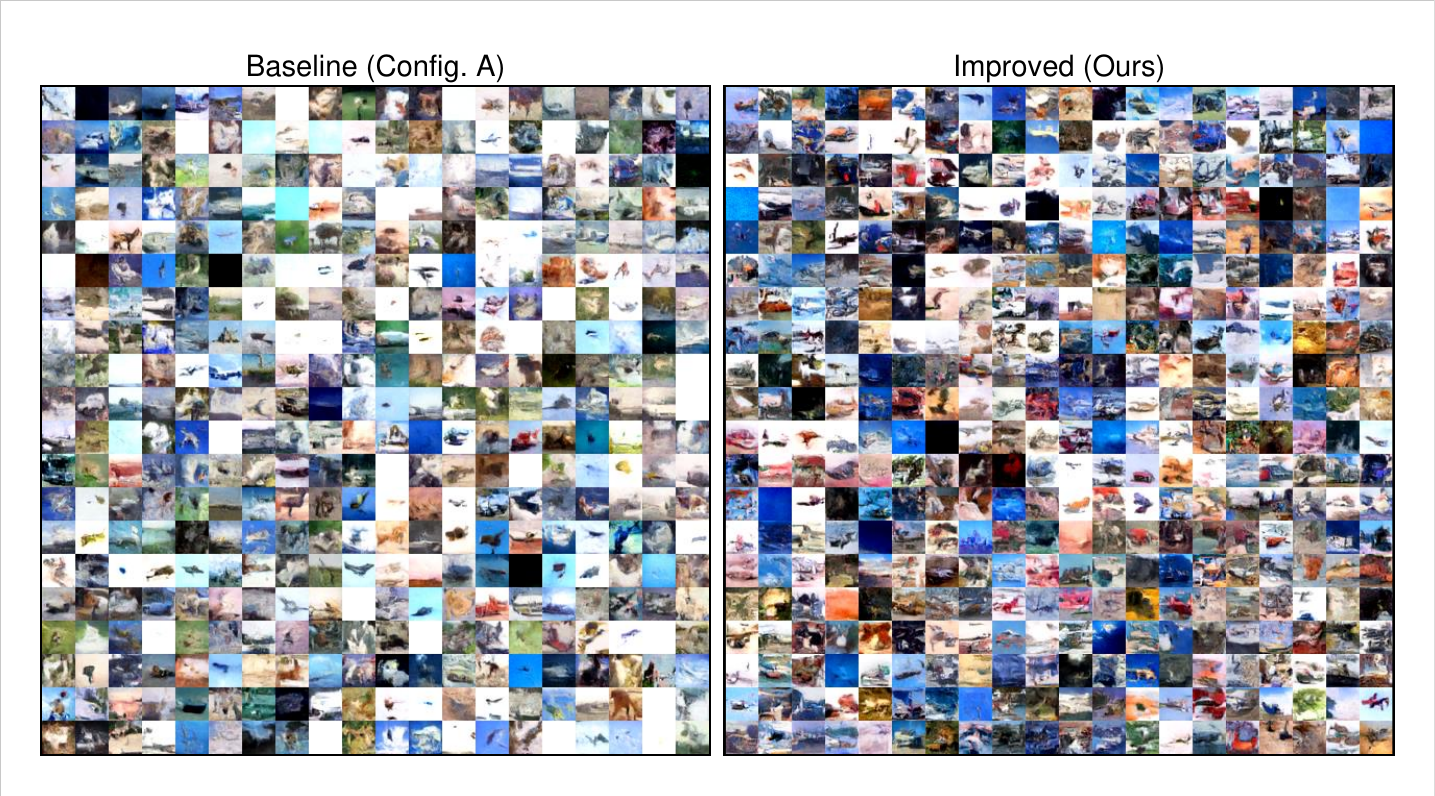}
	\caption{Comparison of generated images by diffusion models trained in CIFAR-10 dataset.}
	\label{Fig2}
\end{figure}

Configuration D-1N (Alias-free resampling and nonlinear enhancements; Kaiser window $\beta=1$, Normalized kernel) achieved the best overall performance for the CIFAR dataset (Table \ref{table_perf}; Figure \ref{Fig2}), significantly outperforming the baseline (Config A) with a FID of 90.21, representing an 8.7\% improvement over the baseline FID and a KID of 5.54 (7.2\% improvement). While its IS of 4.51 is slightly lower than the baseline (4.54), the improvements in FID and KID suggest superior sample quality. Configuration B-0 (Alias-free resampling; No Kaiser window or kernel normalization) also performed well, achieving a FID of 94.23 (4.6\% improvement), a KID of 5.44 (8.9\% improvement) and the highest IS of 4.71.

\subsubsection{MNIST-M}

On the MNIST-M dataset, configuration D-2N yielded the best results, with the lowest FID of 82.46, representing a 3.0\% improvement over baseline (Table \ref{table_perf}; Supplemental Figure \ref{AFig1}) and the lowest KID of 5.35 (14.1\% improvement over the baseline KID of 6.23). These results highlight D-2N as the strongest performer in generating high-quality samples, though its IS of 3.99 was only modestly better than the baseline of 3.76. Configuration B-2N also demonstrated competitive performance with the highest IS of 4.14, a FID of 88.05 (2.3\% lower than the baseline), while its KID of 5.47 was also superior to the baseline by 12.2\%.

\subsubsection{MNIST}

For the MNIST dataset, the baseline architecture remained the best in terms of sample quality (Table \ref{table_perf}; Supplemental Figure \ref{AFig2}). However, configuration B-0 came close, with a FID of 10.23 (6.4\% higher than the baseline) and a KID of 0.58. Config B-2N, while achieving the highest IS of 2.00, did not outperform the baseline in terms of FID or KID.

\subsubsection{Summary of Best Performances}

In summary, Config D-1N outperformed the baseline in the CIFAR-10 dataset, achieving 8.7\% FID and 7.2\% KID improvement. For the MNIST-M dataset, Config D-2N delivered the best results with 3.0\% FID and a 14.1\% KID improvement. These results highlight that just by introducing alias-free resampling into the UNet network, significant improvements in sample quality can be achieved, particularly in terms of FID and KID, which are critical indicators of generative model performance. Note that the reliability of the performance metrics (IS, FID, and KID) on MNIST data is not well-established as the MNIST dataset contains single-channel gray image data while these metrics are designed for RGB images.

\subsection{Assessing Rotational Equivariance}
\begin{figure}[t]
	\centering
	\includegraphics[width=14cm,height=13cm,clip,keepaspectratio]{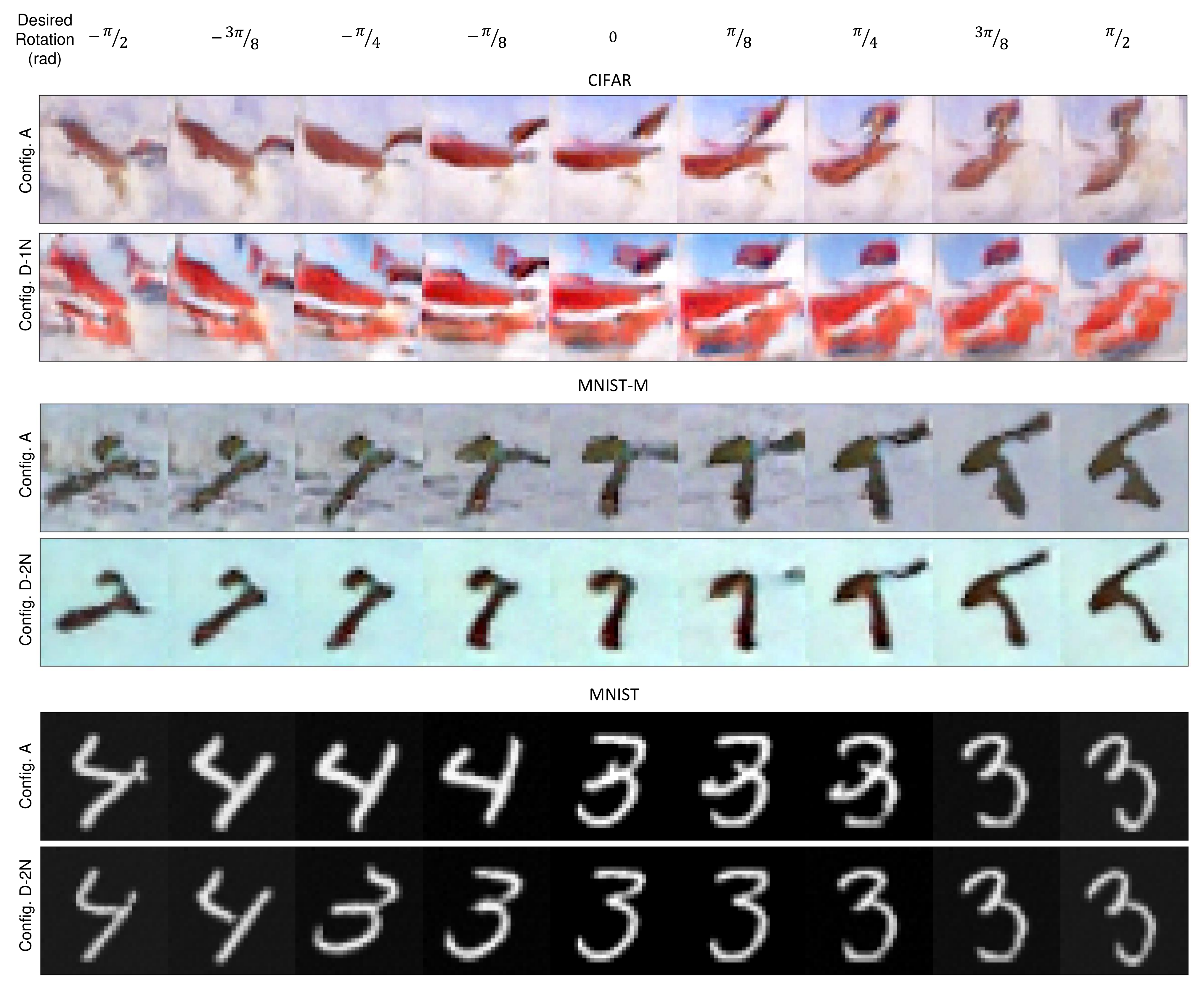}
	\caption{Comparison of generated images by models trained in CIFAR-10 (top), MNIST-M (middle) and MNIST (bottom) dataset with specific desired rotation.}
	\label{Fig3}
\end{figure}

We conducted an initial evaluation to assess the rotational equivariance of our modified diffusion process (Algorithm \ref{alg:sampling2}) by varying the target rotation angle $\phi$ from $-\nicefrac{\pi}{2}$ to $\nicefrac{\pi}{2}$ radians and generating images through the modified diffusion process using models trained on the CIFAR-10, MNIST-M, and MNIST datasets. For each dataset, we compared two models: the baseline (Config. A) and our enhanced UNet (Config. D), with the latter theoretically offering superior robustness to rotation. Figure \ref{Fig3} illustrates the promising ability of our modified diffusion process to generate images at specific rotations without any additional training where our enhanced UNet architecture showed more consistent object rotation for various angles. These results indicate that our additional filtering layers used for alias-free resampling reduce the dependency of image details on absolute pixel coordinates, enabling more coherent image rotation. While these results are encouraging, further comprehensive evaluation is required to confirm its effectiveness.

\subsection{Ablation Studies and Comparative Analysis}
In the ablation study, we analyze the effects of architectural variations (Config B, C, D), the impact of the Kaiser window $\beta$, and the influence of kernel normalization (N) on model performance across the CIFAR-10, MNIST-M, and MNIST datasets by evaluating the contribution of each factor to overall model performance.

\subsubsection{Architecture Comparison (Config B, C, D)}

Across CIFAR-10 and MNIST-M, Config D outperformed Config B and C, particularly in terms of FID and KID, while maintaining competitive IS scores (Table \ref{table_perf}). On CIFAR-10, Config D-1N achieved the lowest FID score, representing an 8.7\% FID and a 7.6\% KID improvement over Config C-2N. Config D-2N also reduced FID by 6.3\% and KID by 20.9\% compared to Config C-2N. On MNIST-M, Config D-2N achieved the best results showing a 17.6\% FID and a 20.7\% KID improvement over Config C-2N. Config B came close with the best KID scores in CIFAR-10 and was particularly superior in MNIST dataset. Indeed on MNIST, Config B-0 achieved the second-best FID and KID while maintaining a superior IS. Overall, Config D provided better FID and KID scores in CIFAR-10 and MNIST-M while Config B achieved the highest IS scores across all datasets with slightly better results in MNIST.

\subsubsection{Effect of Kaiser Window}

No Kaiser window ($\beta = 0$) resulted in higher IS but poorer FID scores (Table \ref{table_perf}). For instance, on CIFAR-10, Config B-0 had the highest IS score, but its FID was worse than Config D-1N by 4.3\%. On MNIST-M, Config B-0's IS score was close to the baseline, but its FID was 13.8\% worse than Config D-2N. 
Kaiser window with $\beta = 1$ provided a more balanced trade-off. For CIFAR-10, Config D-1N achieved the best FID while maintaining strong IS and KID scores. Compared to Config D-0 (no Kaiser window), $\beta = 1$ reduced FID by 7.1\% and KID by 14.8\%.
Finally, Kaiser window with $\beta = 2$ led to the best IS, FID, and KID scores in MNIST-M where Config D-2N achieved a 17.6\% improvement in FID and a 28.1\% improvement in KID over Config D-0.

\subsubsection{Effect of Kernel Normalization}

The introduction of kernel normalization had varying effects on model performance across the datasets. For this, we fixed the Kaiser window $\beta = 1$ and observed the effect of kernel normalization across Config B, C, and D. Configurations with normalization generally improved FID and KID. 
Particularly, on CIFAR, Config D-1N improved IS by 4.40\%, FID by 16.51\%, and KID by 25.34\% compared to Config D-1. Config C-1N also provided a 12.08\% increase in IS, a 22.69\% improvement in FID, and a 33.49\% reduction in KID over Config C-1. 
On MNIST-M, Config B-1N improved IS by 9.12\%, but FID decreased by 7.22\%, and KID was marginally worse by 1.76\%, showing mixed effects. On the other hand, Config C-1N improved IS by 6.03\%, FID by 13.58\%, and KID by 18.88\% over Config C-1. For Config D-1N, IS increased by 6.98\%, while FID improved by 5.36\%, and KID saw an 8.38\% reduction. 
Finally, on MNIST, kernel normalization had no significant effect on IS for Config B and C but improved FID by 5.64\% and KID by 12.5\% for Config B-1N. Config C-1N improved FID by 8.06\% and KID by 17.14\%. Config D-1N saw no IS change, but FID improved by 11.39\%, and KID by 13.39\%, indicating consistent improvements in quality.

\section{Discussion and Future Directions}\label{sec:limitation}

\subsection{Broader Implications}

In this work, we hypothesize that the current resampling operations (upsampling and downsampling) in diffusion model architectures introduce aliasing, which degrades image quality and utilizing theory-driven alias-free resampling can enhance model performance in image synthesis. We proposed architectural modifications of the classical diffusion models by incorporating alias-free resampling into the UNet structure. We demonstrated that our proposed modifications can substantially enhance image quality and model stability without adding complexity or increasing the number of trainable parameters. Our approach aligns with a growing trend in machine learning to leverage domain-specific theories such as signal processing and statistical physics to drive innovation in deep learning architectures \cite{bahri2020statistical,stylegan3}. Furthermore, we proposed a modification of the diffusion process that enables user-controlled image rotation without any additional model training. As computational resources continue to be a bottleneck for such generative models, our work offers a promising theory-driven pathway for achieving customizable high performance in generative modeling without significant increases in computational cost. Finally, our work has the potential to benefit other image-based deep learning architectures by enhancing conventional resampling and nonlinear operations with alias-free resampling techniques. 

\subsection{Limitations}
A primary limitation of this study is the absence of large-scale training on full high-resolution datasets, a constraint imposed by limited computational resources. However, as a proof-of-concept, our goal was to demonstrate the benefits of integrating alias-free resampling via theory-driven modifications of diffusion models for image generation, rather than achieving state-of-the-art results through exhaustive training. To this end, our findings suggest that even under our constrained training conditions, the proposed models consistently outperformed the baseline models, indicating the potential for even greater performance enhancements in large-scale settings. Future work will aim to explore this potential through comprehensive training with full datasets.

\subsection{Future Work}
In this study, we explore the integration of theory-driven alias-free resampling techniques in diffusion models for image generation. Our future works will focus on several avenues to further refine and extend our approach. First, it might interesting to conduct large-scale training on diverse datasets with high-resolution images to fully realize the potential of our proposed alias-free resampling and controlled rotation techniques. Second, we aim to incorporate these techniques into video-generating diffusion models\cite{videocrafter2,ho2022video} to explore their effectiveness in enhancing temporal stability and image coherence across frames. Another key area for future research is a more rigorous assessment of rotational equivariance of controlled rotation to ensure that model performance remains consistent under various rotational transformations. Finally, in this work we kept the filter cutoff and kernel length fixed. However, in future study we plan to investigate filtering with various kernel lengths and cutoffs across different layers of the UNet to enhance image fidelity and stability during generative tasks.

\section{Conclusion}\label{sec:conclude}
In conclusion, this work presents a significant step toward enhancing diffusion models through the integration of alias-free resampling techniques. We hypothesize that the current upsampling and downsampling operations in diffusion model architectures introduce aliasing, which diminishes image quality. We propose modifications to the diffusion model by introducing alias-free resampling within the UNet architecture without adding trainable parameters. Our experimental results across benchmark datasets indicate that these modifications yield consistent quality improvements, particularly in terms of FID and KID, underscoring the effectiveness of theory-driven architectural refinements. Our work not only advances the capabilities of diffusion models but also illustrates the broader potential of incorporating alias-free resampling into other deep learning architectures to achieve efficiency and performance gains. Additionally, we introduce a modified diffusion process with user-controlled rotation, which further demonstrates the potential for more customizable image synthesis. As generative modeling continues to advance, our work offers a pathway for future innovations that provide theory-driven computationally efficient generative architectures.
 
\subsection*{Data and Code Availability}
The actual datasets utilized in this study can be found via this Dropbox \href{https://www.dropbox.com/scl/fi/ll19yhimdi1jscbft7ttm/Diffusion-Model-Datasets.zip?rlkey=d6ahl9ry5brxd9or7rz1emugm&st=a8n19949&dl=0}{link} and the codebase with examples are given in  \href{https://mdfahimanjum.github.io/AliasFree-Diffusion-Models-PyTorch/}{https://mdfahimanjum.github.io/AliasFree-Diffusion-Models-PyTorch/}.


\bibliographystyle{plain} 
\setlength{\itemindent}{0pt} 

\appendix
\section{Appendix}
\begin{figure}[htbp]
	\centering
	\includegraphics[width=13.5cm,height=13cm,clip,keepaspectratio]{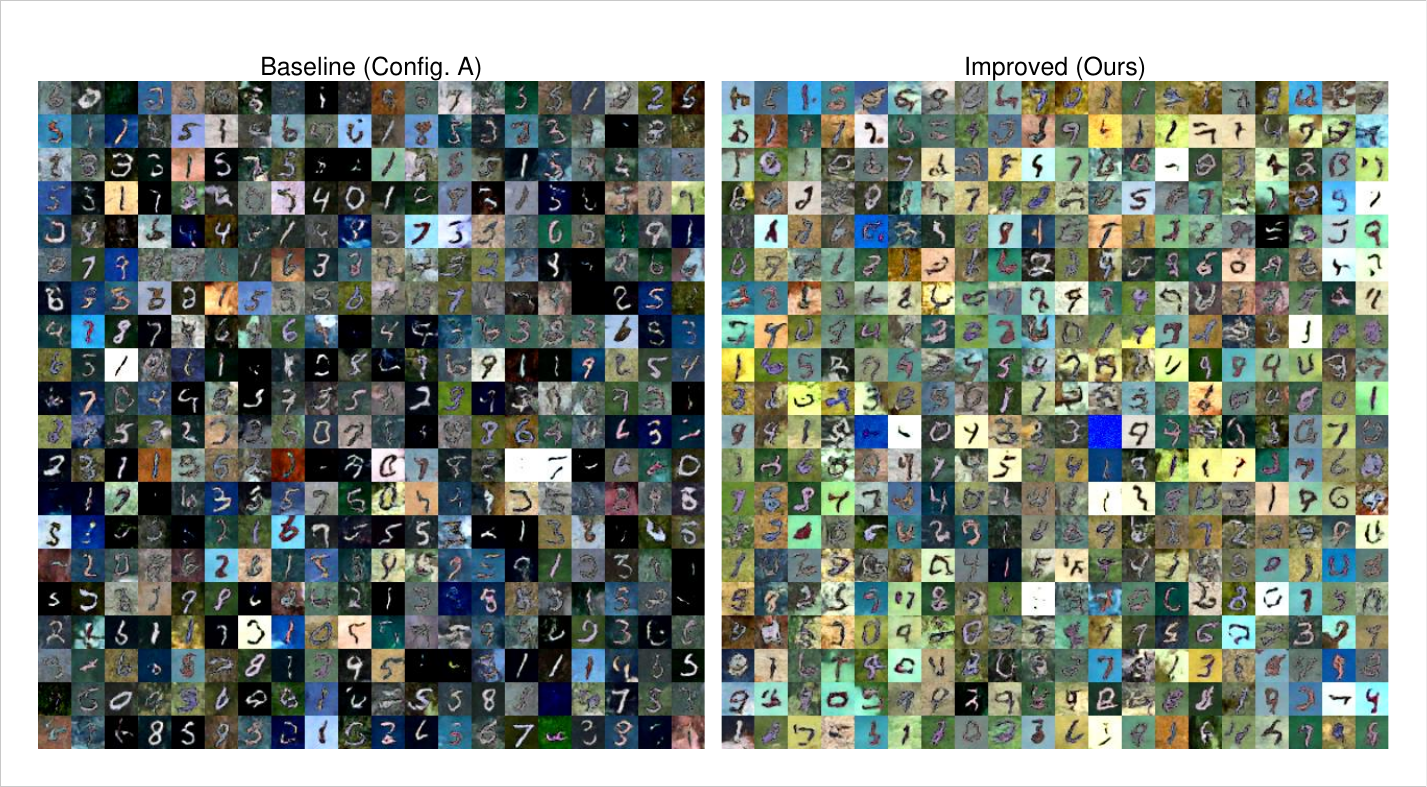}
	\caption{Comparison of generated images by diffusion models trained in MNIST-M dataset.}
	\label{AFig1}
\end{figure} 
\begin{figure}[htbp]
	\centering
	\includegraphics[width=13.5cm,height=13cm,clip,keepaspectratio]{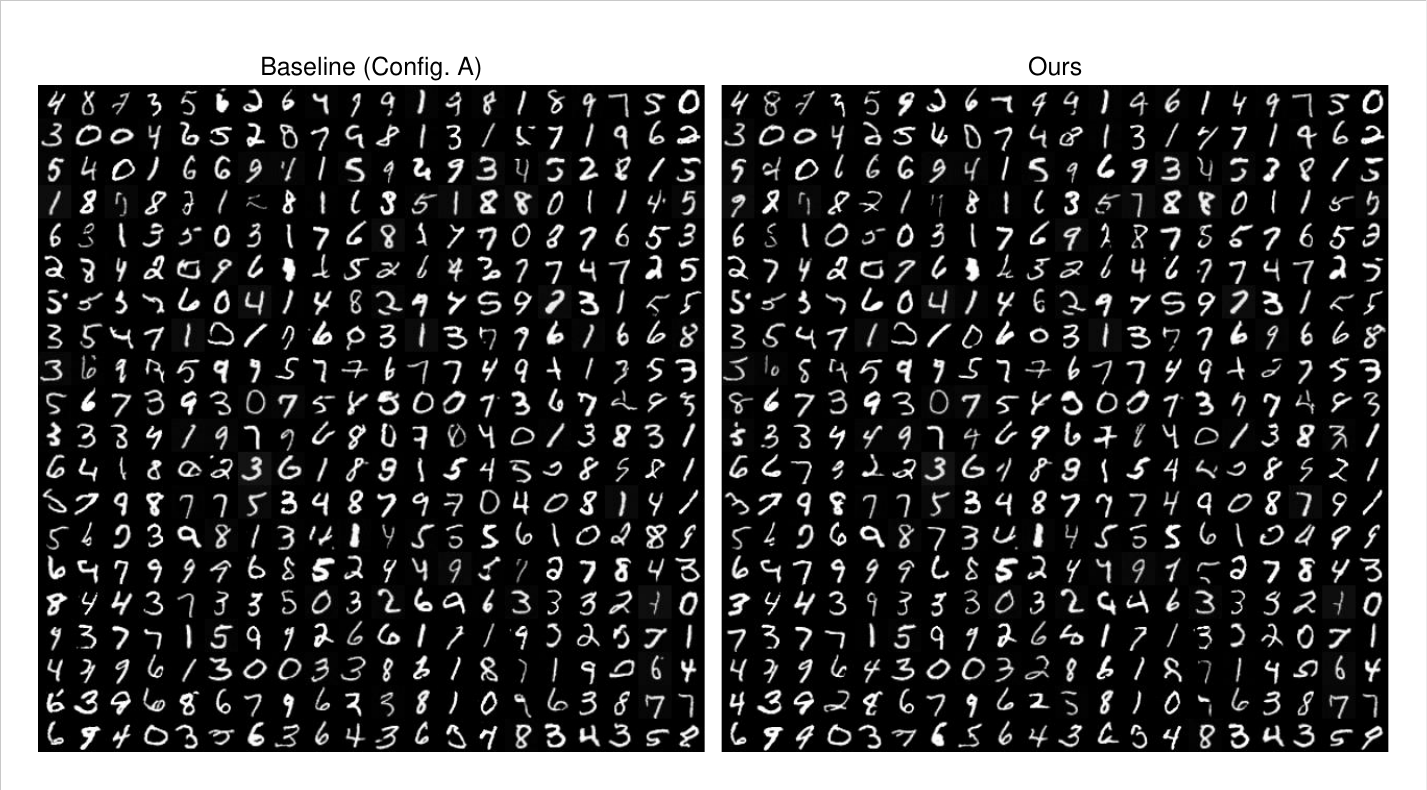}
	\caption{Comparison of images generated by diffusion models trained on the MNIST dataset.}
	\label{AFig2}
\end{figure}
\vfil   
\end{document}